\documentclass{svproc}
\usepackage{url}
\usepackage{lineno,hyperref}
\usepackage{algorithm,algpseudocode}
\usepackage{multirow}
\usepackage{graphicx}
\usepackage[font=small,labelfont=bf]{caption}
\modulolinenumbers[5]

\begin{document}
\mainmatter              
\title{An approach to extract information from academic transcripts of HUST}
\titlerunning{An approach to extract information from scoreboards of HUST}  
%
\author{Nguyen Quang Hieu\inst{1a}, Nguyen Le Quy Duong\inst{2}, Le Quang Hoa\inst{1b} (*), Nguyen Quang Dat\inst{3}}
\authorrunning{N.Q. Hieu, N.L.Q. Duong, L.Q. Hoa, N.Q. Dat} 
%
\tocauthor{N.Q. Hieu, N.H. Khang, L.Q. Hoa, N.Q. Dat}
\institute{School of Applied Math. and Informatics, Hanoi University of Science and technology, Hanoi, Vietnam\\
(a) \textit{hieu.nq185351@sis.hust.edu.vn};\\
(b) \textit{hoa.lequang1@hust.edu.vn}
\and
School of Information and Communications Technology, Hanoi University of Science and technology, Hanoi, Vietnam\\
\textit{duong.nlq210242@sis.hust.edu.vn}
\and
HUS High School for Gifted Students, Hanoi University of Science, Vietnam National University, Hanoi, Vietnam\\
\textit{nguyenquangdat@hus.edu.vn}\\
(*) Corresponding author.}

\maketitle              

\begin{abstract}
In many Vietnamese schools, grades are still being inputted into the database manually, which is not only inefficient but also prone to human error. Thus, the automation of this process is highly necessary, which can only be achieved if we can extract information from academic transcripts. In this paper, we test our improved CRNN model in extracting information from 126 transcripts, with 1008 vertical lines, 3859 horizontal lines, and 2139 handwritten test scores. Then, this model is compared to the Baseline model. The results show that our model significantly outperforms the Baseline model with an accuracy of 99.6\% in recognizing vertical lines, 100\% in recognizing horizontal lines, and 96.11\% in recognizing handwritten test scores.
\end{abstract}

\section{Introduction}

\quad

At Hanoi University of Science \& Technology, after every exam, scores are recorded in academic transcripts and then transferred to the school's database by teachers. Until now, this process has been done manually, which is time-consuming for the teachers and may lead to accidental mistakes such as the scores inputted being incorrect or the scores being assigned to the wrong students.

Currently, machine learning methods have been applied to automate these processes (see \cite{Vu2022}, \cite{Anh2022}, \cite{Hoang2022}, \cite{Williams1986}, \cite {CuKimLong}, \cite{NguyenXuanTung}, \cite{PVHai1}, \cite{PVHai2}, \cite{LongCK}, \cite{DungLe}). It also helps to free up manpower. By utilizing Image-Processing Techniques and Deep Learning, we can automate this procedure with a system that can extract necessary data.

This paper contains five sections. The introduction is the first section. The second section consists of prior research conducted by various authors on image processing research techniques. The third part is the method studied in this paper, including our proposed method. The fourth part is our results based on real data. And the last part is conclusion and acknowledgment.

\section{Related works}

\quad 

Digit recognition is extremely useful. This is especially the case for schools where the process of inputting grades into database is still being done manually. In such schools, the assistance of digit recognition can significantly increase accuracy and reduce the time allotted to this process.

In 2018, Liu et al. \cite{Liu2018} proposed a hybrid CNN-LSTM algorithm for identifying CO$2$ welding defects. The algorithm is evaluated using 500 images of molten pools from the Robotics and Welding Technology Laboratory at Guilin University of Aerospace Technology. The results from CNN-LSTM algorithm were considered to be better than those of other basic algorithms (CNN, LSTM, CNN-3), with an accuracy of 85\%, 88\%, and 80\% for 32x32, 64x64, and 128x128 images, respectively.

In 2019, Rehman et al. \cite{Rehman2019} utilized a hybrid CNN-LSTM model to analyze opinions in people's online comments. This model outperforms conventional machine learning techniques on the IMDB movie reviews dataset and the Amazon movie reviews dataset, with a precision of 91\%, recall of 86\%, an F-measure of 88\%, and an accuracy of 91\%.

In 2020, Yang et al. \cite{Yang2020} compared the performance of CNN-LSTM model with analytical analysis and FEA algorithm in detecting the natural frequency of six types of beams. The goal was to assess the first, second, and third-order frequencies of each beam. Author's model was concluded to be superior in both the test on robustness, with 96.6\%, 93.7\%, and 95.1\% accuracy, respectively, and the test on extrapolability, with 95.4\%, 92\%, and 92.5\% accuracy, respectively.

In 2019, Sivaram et al. \cite{Sivaram2019} proposed a CNN-RNN combination model to detect facial landmarks. The proposed model outperforms existing methods, such as CFAN, SDM, TCDN, and RCPR, on the FaceWarehouse database, with 99\% precision, 99\% recall, 99\% F1-Score, 98.65\% Accuracy, and 98.65\% AUC or ROC.

In 2018, Xiangyang et al. \cite{Xiangyang2018} utilized a hybrid CNN-LSTM model to tackle the problem of text classification. With the Chinese news dataset (proposed by the Sogou Lab), the model proved superior to traditional KNN, SVM, CNN, and LSTM, with a precision of 90.13\% under the CBOW model and 90.68\% under the Skip-Gram model.

In 2017, Yin et al. \cite{Yin2017} used CNN-RNN and C3D hybrid networks to detect emotions in videos from the AFEW 6.0 database. The objective was to assign emotion from 7 categories, namely anger, disgust, fear, happiness, sadness, surprise, and neutral, to each video in the test dataset. It was found that with 1 CNN-RNN model and 3 C3D models, an accuracy of 59.02\% was achieved, surpassing the baseline accuracy of 40.47\% and last year's highest accuracy of 53.8\%.

In 2017, Zhan et el. \cite{Zhan2017} introduced a new RNN-CTC model to recognize digit sequences in three datasets, namely CVL HDS, ORAND-CAR (include CAR-A and CAR-B), and G-Captcha. Even though the proposed model only achieved a recognition rate of 27.07\% for the CVL dataset, the model outperformed other state-of-the-art methods on the CAR-A, CAR-B, and G-Captcha datasets, with recognition rates of 89.75\%, 91.14\%, and 95.15\%, respectively, due to the absence of segmentation.

\section{Methodology}

\subsection{Convolutional Neural Networks}

\quad 

A convolutional neural network can be successfully applied for most computer vision problems. Its characteristics make it more effective than other conventional methods. Since its inception, CNN has witnessed numerous optimizations.

However, when it comes to deeper networks, a degradation problem arises. To tackle this, He et al. proposed a deep residual learning framework, ResNet. The basic structure of this network is using shortcut connections. Shortcut connections are those that traverse multiple layers. With this type of connection, we can overcome the problem of vanishing gradients and construct more extensive networks; in other words, better feature representations can be acquired. In practice, shortcut connections can be adjusted on a case-by-case basis, depending on each specific problem.

In our proposed model, we design a 10-layer residual network that doesn't have any global pooling layers. To prevent divergence, we avoid employing CNN that is excessively profound. In addition, we maximize the use of residual learning to enhance gradient propagation.

\subsection{Recurrent Neural Networks}

\quad 

A Recurrent neural network is an architecture where multiple connections between neurons create a recursive cycle. Self-connection brings the advantage of utilizing contextual data when making a correspondence between sequences of input and output. Nevertheless, for conventional RNN, the range of accessible context is restricted in practice. due to the vanishing gradient problem. Applying a memory structure to RNN, which produces a cell that is known as a long short-term memory (LSTM), is one method that can be utilized to solve the problem. It has been demonstrated that the LSTM variant of RNN is capable of resolving some of the issues that are inherently present in conventional RNN, in addition to learning how to resolve issues of long-term dependency. At this time, LSTM has developed into one of the RNNs that is utilized the most frequently.

Regarding the sequence labeling problem, it is helpful to have access to context in both the future and the past. However, the normal LSTM solely considers information from the past and pays no attention to the context of the future. The creation of a second LSTM that processes input in the other direction is one alternate solution. This type of LSTM is known as a bidirectional LSTM, and its abbreviation is Bi-LSTM. Every training sequence is conveyed both forward and backward to two distinct LSTM layers, both of which are connected to the same output layer by the Bi-LSTM algorithm. This structure provides the output layer with full context for all points in the input sequence, both in the past and in the future. This context is supplied throughout the entire time period.

\subsection{Handwritten digit string recognition with CTC}

    \quad 
    Sequence characters recognition is a common problem of OCR. In this paper, we proposed an approach to recognize handwritten digit string. After having features extracted by a convolutional neural network, the main idea is to use an output connectionist temporary classification layer to get the final predicted results after using a recurrent neural network to recognize sequence information in images. This is done after the convolutional neural network has been used to extract features.
    
    The input image is one-dimensional tensor (after resizing 40x100x1). For feature extraction, a backbone network is constructed with convolutional, max-pooling layers, and residual network.  After every convolution layer, we performed batch normalization to prevent internal covariance shift. Output of feature extraction block are fed as a sequence into labeling block. To advoid vanishing gradient problem, we use two Bi-LSTM layers instead of traditional RNN layer. Finally, a fully connected layer is used to reduce the dimension to 13, before passing CTC layer. The CTC layer serves primarily two functions: the first of these is to decode the output, and the second is to calculate the loss. The full architecture is shown in figure \ref{fig:architecture}
    \begin{center}
        \includegraphics[width=0.7\linewidth]{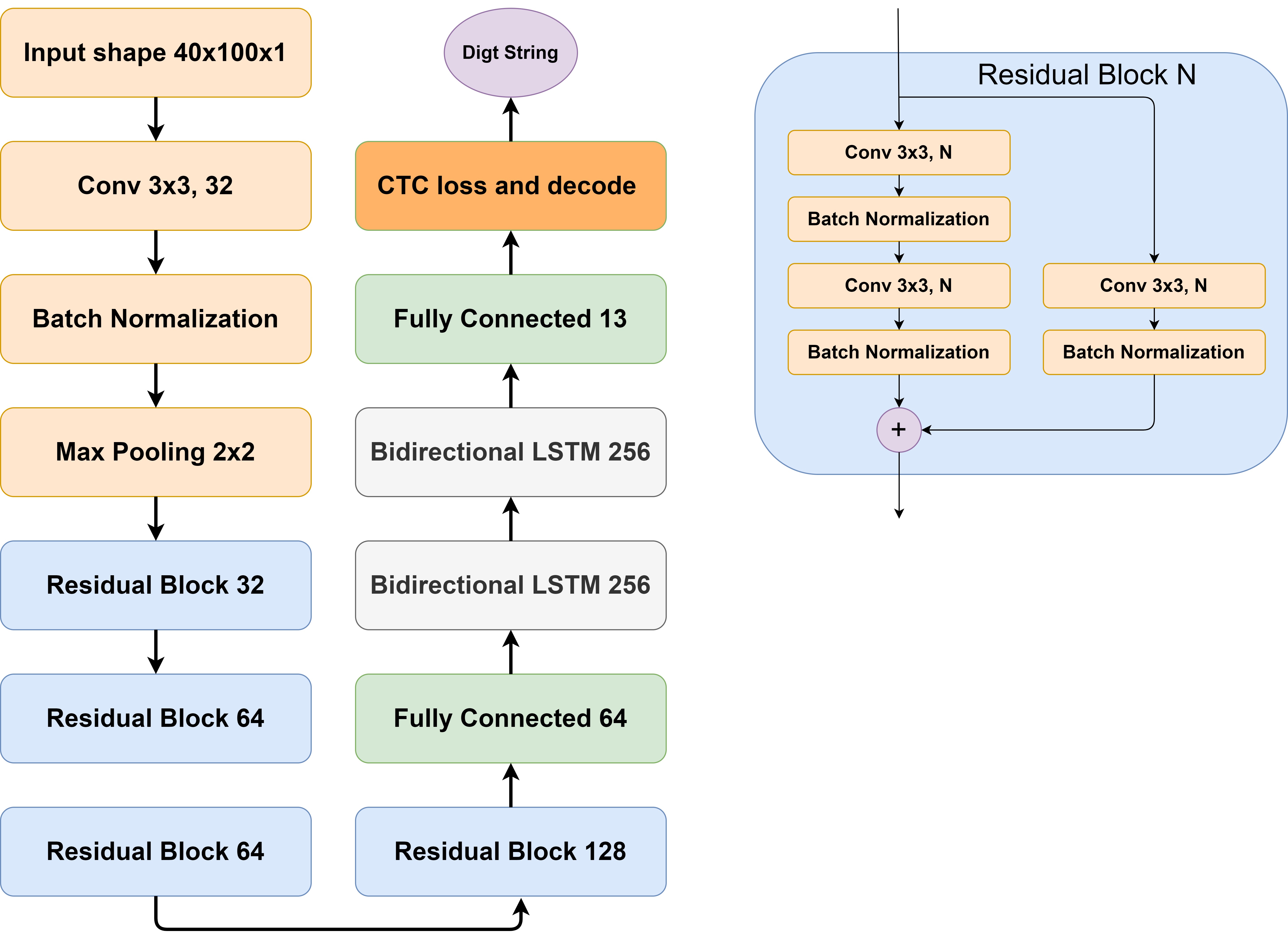}
		\captionof{figure}{Our proposed architecture model}
		\label{fig:architecture}
	\end{center}

\subsection{Proposed method }
	
	\begin{center}
	    \begin{figure}[!htb]
				\centering
				\includegraphics[width=\linewidth]{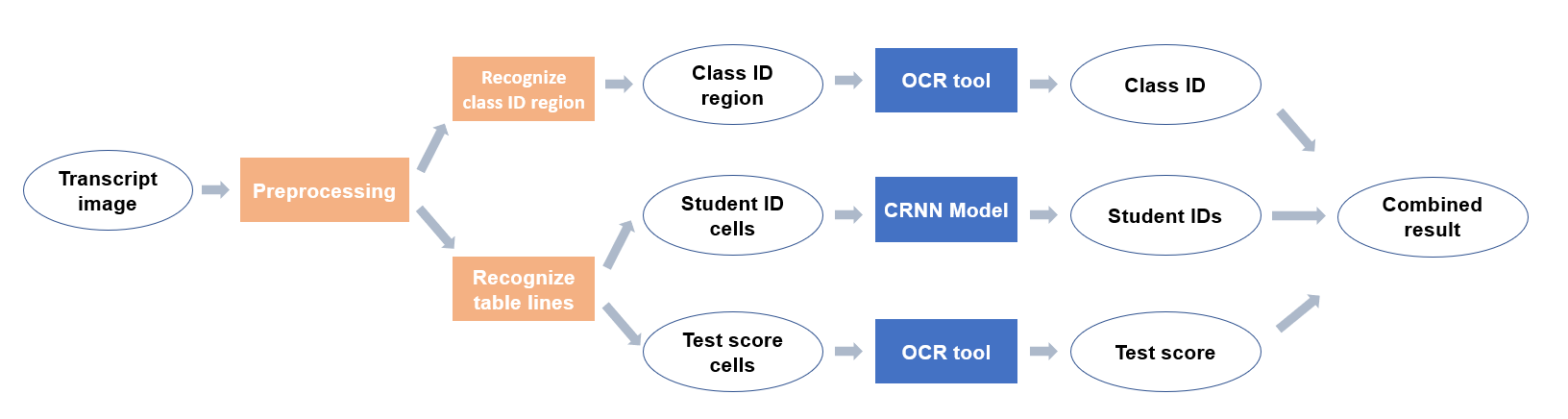}
				\captionof{figure}{Our proposed method flowchart  }
		\end{figure}
	\end{center}
	
    The first step of our method is image preprocessing. Transcript images are binarize, removing noises by Gaussian filter. We deskew the images by Projection profile algorithm. For class ID recognition, we use Template matching followed by OCR tools. To recognize and calculate coordinates of lines in transcript, horizontal and vertical masks generated by morphological operations are fed into Hough transformation. After having full coordinates of lines, cells of student IDs and test scores are cropped. For student IDs, are sequence of printed digts, can easily recognized by available OCR tools. In our method, we use Tesseract-OCR, which is inherently very accurate in recognizing printed characters. For test scores, we use our Handwritten digits recognition model with CTC decode. Finally, student IDs and test scores are combined.  
    \begin{center}
		\begin{figure}[!htb]
				\centering
				\includegraphics[width=\linewidth]{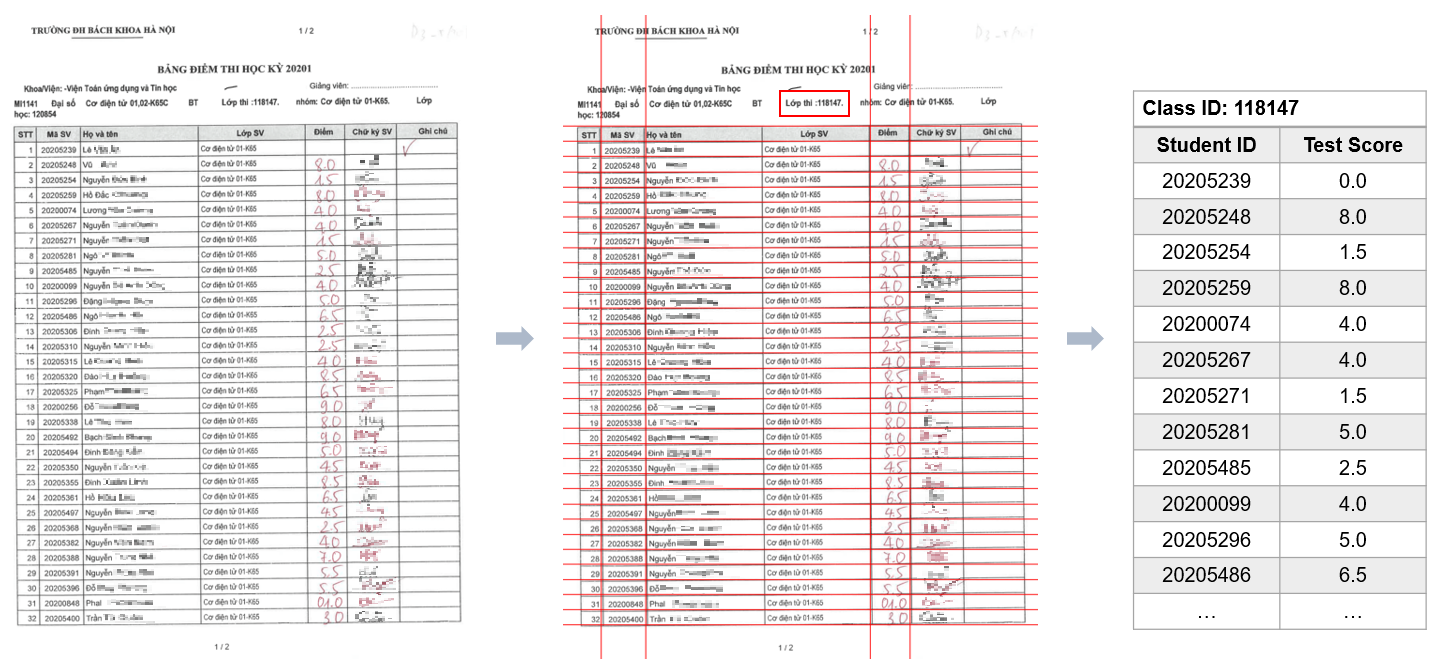}
				\captionof{figure}{Result of automatic score-inputting system }
	    \end{figure}
	\end{center}

\section{Experiment and results}\label{Exp}

    \subsection{Evaluation of Image-preprocessing}
	By using a dataset consisting of images of 126 academic transcripts with 1008 vertical lines and 3859 horizontal lines, the results of the Baseline model and my improved model in detecting lines are shown below:

	\begin{center}
		\includegraphics[width=8cm]{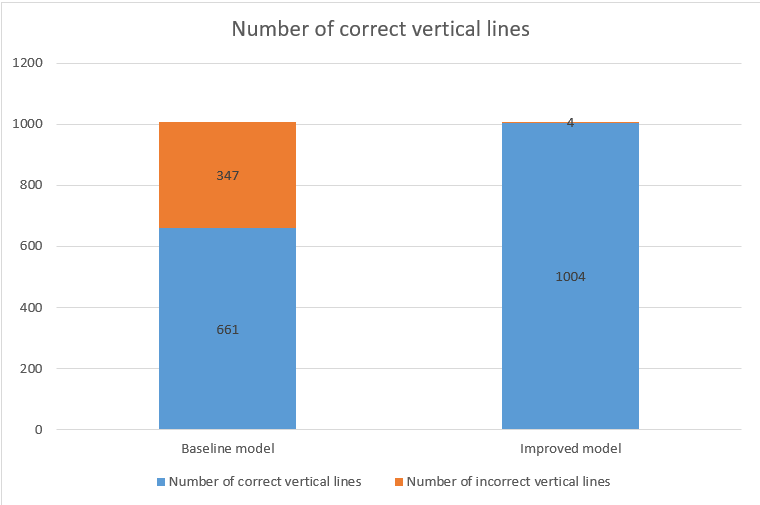}
		\captionof{figure}{Results of the two models in detecting vertical lines}
	\end{center}
	The Baseline model achieved an accuracy of 65.57\% for vertical lines, whereas the improved model had an accuracy of 99.6\%.
	
	\begin{center}
		\includegraphics[width=8cm]{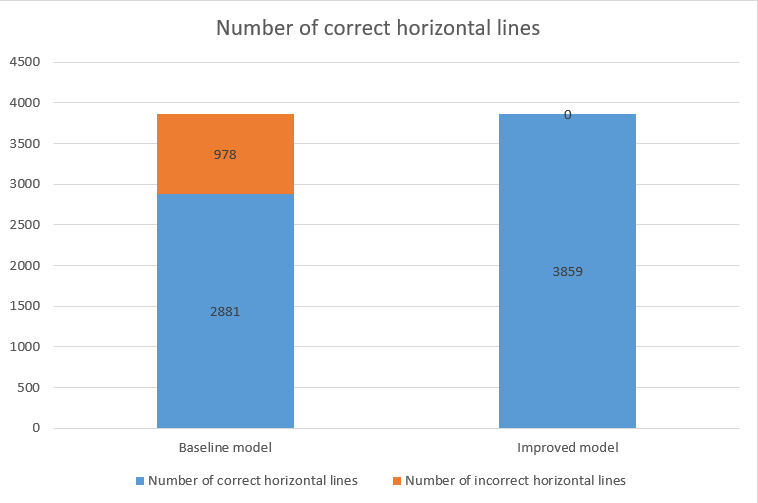}
		\captionof{figure}{Results of the two models in detecting horizontal lines}
	\end{center}
	The Baseline model achieved an accuracy of 74.65\% for horizontal lines, whereas the improved model had an absolute accuracy of 100\%.
	
	\subsection{Evaluation of models in recognizing handwritten test scores}
	By using an extracted dataset with 2139 handwritten test scores, the results of the CNN Baseline model and the my CRNN model are shown below:
	
	\begin{center}
		\includegraphics[width=8cm]{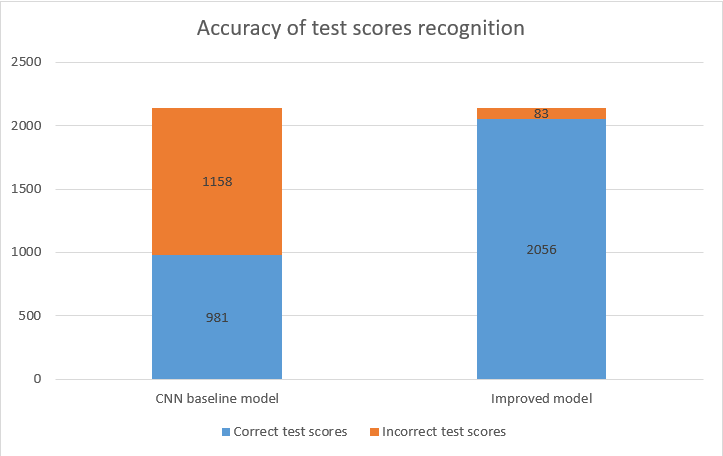}
		\captionof{figure}{Results of the two models in recognizing handwritten test scores}
	\end{center}
	The Baseline model achieved an accuracy of approximately 45.86\%, whereas my improved CRNN model had an accuracy of 96.11\%.

	\subsection{Evaluation of automatic score-inputting system}
	To evaluate the accuracy of the entire score-inputting system, we tested it on a dataset of 75 scanned academic transcripts with 162 images of size 1653 x 2337.
	
	\subsubsection{a) Evaluation of Baseline model\\}
	Results of the Baseline model:
	\begin{itemize}
		\item The model was able to accurately detect lines in 92 images and recognize class IDs of 51 images (20 academic transcripts), achieving an accuracy of 31.4\%.
		\item Among 3596 student IDs, the model correctly recognized 2201 IDs, achieving an accuracy of 61.2\% (The majority of images in which the lines were accurately detected all had their student IDs recognized by the model).
		\item Among 3596 test scores, the model was able to accurately recognize 1532 test scores, achieving an accuracy of 42.6\%.
	\end{itemize}
	
	\subsubsection{b) Evaluation of improved model\\}
	Results of the improved model:
	\begin{itemize}
		\item In 22 images, the model misidentified 1 vertical line. However, these misidentified lines didn't affect the columns of data that needed to be recognized. Horizontal lines, on the other hand, were all accurately detected. In all 162 images, the model correctly recognized the class IDs with an accuracy of 100\%. 
		\item Among 3596 student IDs, the model correctly recognized 3481 IDs, achieving an accuracy of 96.8\%.
		\item Among 3596 test scores, the model was able to accurately recognize 3451 test scores, achieving an accuracy of 95.9\%.
	\end{itemize}

\section{Conclusion}

\quad 

In this research paper, we have introduced a new approach to the case of handwritten test scores into the computer. When using additional auxiliary features on the printout such as horizontal and vertical lines on the A4 paper, we have achieved very good results in clearly separating handwritten letters and numbers, thereby increasing adds precision to reading handwritten letters and numbers into numeric data.

In the future, we will put more research on some related issues, in order to further increase the accuracy:

\quad - Identify several records of the same person.

\quad - Identify both letters and numbers at the same time (points are written in both numbers and words in the one handwritten paper)

\section{Acknowledgment}

\quad 

We would like to extend our gratitude to the researchers at Hanoi University of Science and Technology (Hanoi, Vietnam), who were of great assistance to us in the process of collecting data and testing the application of the model in real-world scenarios.

%
%

\end{document}